\documentclass[10pt,twocolumn,letterpaper]{article}

\usepackage{iccv}              

\definecolor{iccvblue}{rgb}{0.21,0.49,0.74}
\usepackage[pagebackref,breaklinks,colorlinks,allcolors=iccvblue]{hyperref}

\usepackage{times}
\usepackage{latexsym}
\usepackage{amsmath}
\usepackage{graphicx}
\usepackage{subcaption}
\usepackage{booktabs}
\usepackage{xcolor}
\usepackage{colortbl}
\usepackage{multirow}
\usepackage{balance}
\usepackage{longtable}
\usepackage{tcolorbox}
\usepackage{array}

\usepackage[T1]{fontenc}

\usepackage[utf8]{inputenc}

\usepackage{microtype}

\usepackage{inconsolata}

\usepackage{graphicx}
\usepackage{float}

\usepackage{colortbl}     
\usepackage{arydshln} 

\def\paperID{*****} 
\def\confName{ICCV}
\def\confYear{2025}

\title{CountQA: How Well Do MLLMs Count in the Wild?}

\author{
    Jayant Sravan Tamarapalli\thanks{Equal Contribution} \\
    Google\\
    {\tt\small jayantsravan@google.com}
\and
    Rynaa Grover\footnotemark[1] \\
    Google\\
    {\tt\small rynaa@google.com}
\and
    Nilay Pande\footnotemark[1] \\
    Waymo\\
    {\tt\small nilayp@waymo.com}
\and
    Sahiti Yerramilli\footnotemark[1] \\
    Google\\
    {\tt\small sahitiy@google.com}
}
\makeatletter
\renewcommand\@fnsymbol[1]{\ensuremath{\ifcase#1\or \dagger\or \ddagger\or \mathsection\or \mathparagraph\or \|\or **\or \dagger\dagger\or \ddagger\ddagger\else\@ctrerr\fi}}
\makeatother

\begin{document}
\maketitle
\begin{figure*}[t!]
\centering
\includegraphics[width=\textwidth]{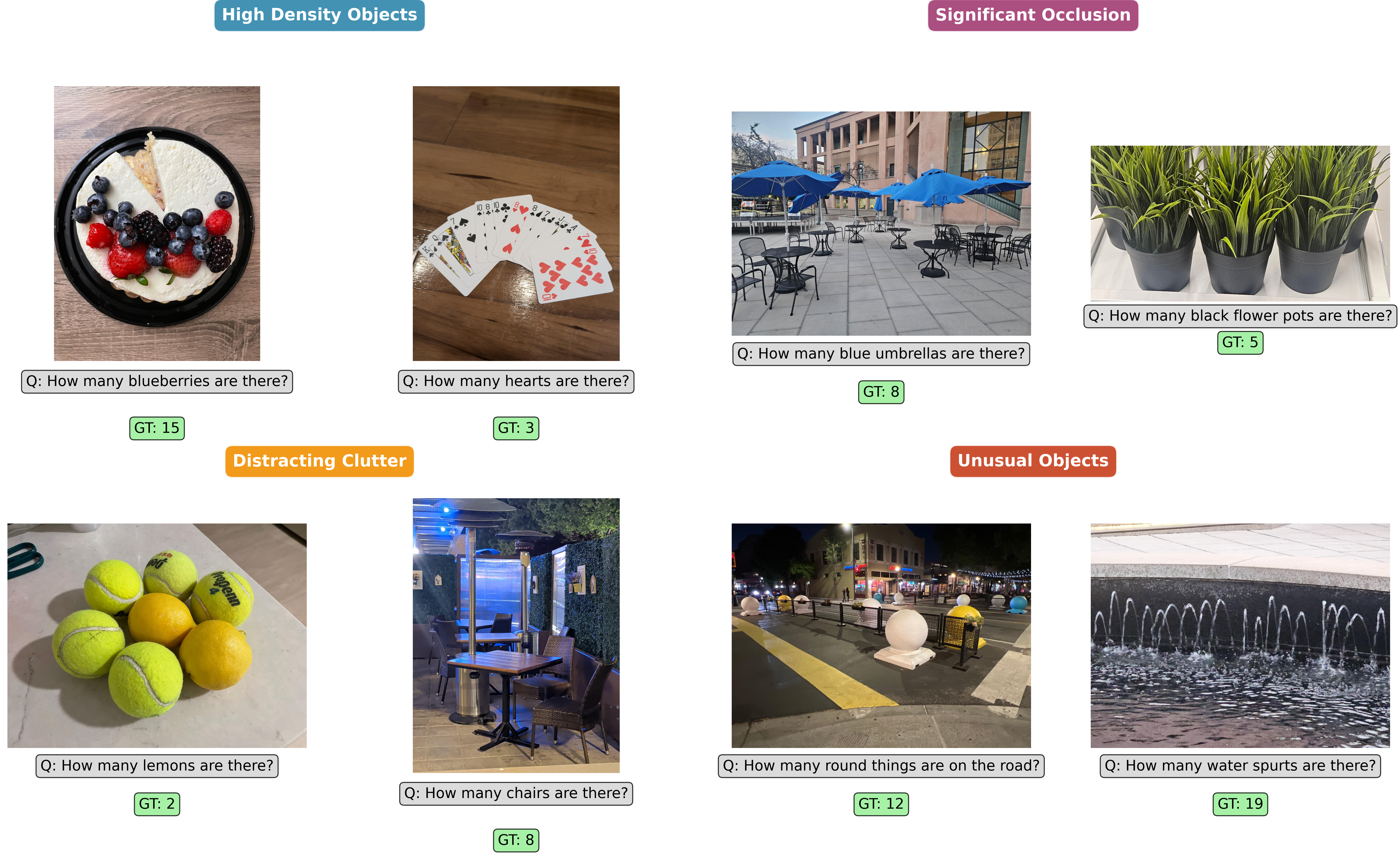}
\caption{An overview of our new dataset, \textbf{CountQA}, designed to challenge the counting abilities of MLLMs with difficult, real-world scenes. Our dataset features scenes with \textbf{high object density}, significant \textbf{visual clutter}, \textbf{occlusion}, and \textbf{unusual objects} not typically found in standard counting benchmarks. Each image is paired with a natural language question and a meticulously annotated ground truth (GT) count.}
\label{fig:teaser}
\end{figure*}

\begin{abstract}
Multimodal Large Language Models (MLLMs) demonstrate remarkable fluency in understanding visual scenes, yet they exhibit a critical lack in a fundamental cognitive skill: object counting. This blind spot severely limits their reliability in real-world applications. To date, this capability has been largely unevaluated in complex scenarios, as existing benchmarks either feature sparse object densities or are confined to specific visual domains, failing to test models under realistic conditions. Addressing this gap, we introduce \textbf{CountQA}, a challenging new benchmark designed to probe this deficiency. Comprising over 1,500 question-answer pairs, CountQA features real-world images with high object density, clutter, and occlusion. We investigate this weakness by evaluating 15 prominent MLLMs on the CountQA benchmark and reveal that the top-performing model achieves a mere 42.9\% accuracy, with performance declining as object counts rise. By providing a dedicated benchmark to diagnose and rectify this core weakness, CountQA paves the way for a new generation of MLLMs that are not only descriptively fluent but also numerically grounded and spatially aware. We will open-source the dataset and code upon paper acceptance to foster further research.

\textbf{Dataset:} \href{https://huggingface.co/datasets/Jayant-Sravan/CountQA}{CountQA on Hugging Face}

\end{abstract}    
\section{Introduction} \label{sec:introduction}

The advent of Multimodal Large Language Models (MLLMs) represents a landmark achievement in artificial intelligence, bridging the gap between human language and complex visual data. Models demonstrate a stunning array of capabilities, from generating rich descriptions of images to engaging in sophisticated, visually-grounded reasoning \cite{geminiteam2025geminifamilyhighlycapable, comanici2025gemini25pushingfrontier, gemmateam2025gemma3technicalreport, openai2024gpt4technicalreport, llava, anthropic2025claude, bai2025qwen25vltechnicalreport, jiang2023mistral7b, agrawal2024pixtral12b}. The advanced capabilities of these models give the appearance of a comprehensive, human-like understanding.

However, beneath this impression of fluency lies a surprising and critical fragility in their fundamental cognitive skills. Specifically, these powerful models exhibit a profound and brittle limitation in the elementary task of object counting \cite{qharabagh2025lvlmcountenhancingcountingability}. This failure calls into question the very nature of their acclaimed visual intelligence. Object counting is not a niche skill, but a cornerstone of visual reasoning. Without it, a model cannot reliably compare quantities, identify a majority, or grasp fundamental concepts like surplus and scarcity. Ultimately, an MLLM that fails at simple enumeration offers only a superficial description of the world, failing to transition from pattern recognition to genuine quantitative understanding. 

While current MLLMs can enumerate small quantities of objects, a skill analogous to human \textit{subitizing}, this ability breaks down dramatically as object counts increase \cite{qharabagh2025lvlmcountenhancingcountingability}. This struggle reveals a fundamental weakness that is less numerical and more perceptual: a failure in fine-grained spatial perception and object individuation. This perceptual deficit is evident in their inability to perform robust part-whole reasoning, for instance, models often require complex 'object-aware' processing to avoid miscounting a single bisected object as two separate entities \cite{qharabagh2025lvlmcountenhancingcountingability}. Corroborating this, other benchmarks show MLLMs also struggle with fine-grained visual grounding in out-of-distribution scenes \cite{grover2025huemanityprobingfinegrainedvisual, li2025countsbenchmarkingobjectdetectors, yerramilli2025geochainmultimodalchainofthoughtgeographic} and even with simpler diagrammatic images \cite{sun2025mathglancemultimodallargelanguage}. Therefore, counting in cluttered, real-world scenes serves as a powerful proxy for probing these critical perceptual limits in modern AI.

This profound shortcoming exposes a blind spot in current evaluation methodologies. Existing benchmarks are ill-equipped for this challenge: traditional datasets target specialized tasks like high-density crowd counting, while modern MLLM evaluations prioritize a broad spectrum of general capabilities. Consequently, a critical question has been largely ignored: can these models truly count?

To fill this gap, we introduce \textbf{CountQA}, a benchmark designed to probe the limits of numerical reasoning and perceptual acuity in MLLMs. Our contributions are threefold:
\begin{enumerate}
    \item \textbf{Challenging Real-World Dataset:} These images, manually collected and annotated, depict everyday indoor and outdoor scenes intentionally selected for their difficulty. As shown in Figure \ref{fig:teaser}, they feature dense arrangements of objects with significant clutter and occlusion to rigorously test the boundaries of MLLM performance.
    \item \textbf{Comprehensive Benchmarking:} We provide a thorough evaluation of 15 prominent MLLMs, including both proprietary, closed-source models and leading open-source alternatives. This establishes strong and extensive baseline performances on our new benchmark.
    \item \textbf{Rich Annotations for Multi-Faceted Analysis:} The dataset includes rich annotations for object classes, image categories, and the presence of distractor elements. This multifaceted annotation scheme enables a detailed, fine-grained analysis of model failure modes.
\end{enumerate}

Our extensive experiments reveal the stark difficulty of this task: the best-performing MLLM achieves an Exact Match accuracy of only 42.9\%, with the next-best model trailing significantly at 34.42\%. By providing the community with a dedicated tool to probe this fundamental limitation, our work aims to catalyze the development of more spatially aware and numerically grounded MLLMs.
\section{Related Works} \label{sec:related}

The field of object counting has evolved from highly specialized systems to flexible, general-purpose models, creating a persistent tension between accuracy in constrained domains and adaptability in open-world scenarios. Our work confronts this frontier by evaluating whether today's adaptable models can handle counting tasks once reserved for specialized systems.

Historically, object counting relied on two main paradigms. Detection-based methods using models like YOLO \cite{redmon2016lookonceunifiedrealtime} and Faster R-CNN \cite{ren2016fasterrcnnrealtimeobject} work well for sparse scenes but fail for scenes with high density and occlusion. To address this, the community largely shifted to density map regression \cite{NIPS2010_fe73f687, 7780439}, which excels in dense crowds but is inherently inflexible to unseen object categories. More recently, the advent of foundation models like CLIP \cite{radford2021learningtransferablevisualmodels} and SAM \cite{kirillov2023segment, ravi2024sam2segmentimages} enabled flexible, ``zero-shot'' counting methods. These approaches, including frameworks like CountGD \cite{amininaieni2025countgdmultimodalopenworldcounting} built on detectors like GroundingDINO \cite{liu2024groundingdinomarryingdino}, represent a monumental leap in flexibility but have yet to demonstrate the fine-grained perceptual acuity required for reliable counting in complex scenes \cite{qian2025t2icountenhancingcrossmodalunderstanding, shi2023trainingfreeobjectcountingprompts}.

This gap is evident in the current evaluation landscape. While canonical high-density datasets like ShanghaiTech \cite{7780439}, UCF-QNRF \cite{idrees2018compositionlosscountingdensity}, JHU-CROWD++ \cite{sindagi2020jhucrowdlargescalecrowdcounting}, and NWPU-Crowd \cite{Wang_2021} were designed for the older, specialized regressors, their narrow focus on a single object category - crowds of people - makes them ill-suited for evaluating the general-purpose counting abilities of modern MLLMs. Conversely, contemporary MLLM benchmarks test related but distinct skills, such as relative quantity comparison \cite{kil2025mllmcompbenchcomparativereasoningbenchmark}, visual grounding prerequisites \cite{li2025countsbenchmarkingobjectdetectors}, or multiple-choice formats \cite{chen2025evaluatingadvancingmultimodallarge, seedbench1}. Similarly, while general VQA datasets like VQA v2 \cite{goyal2017makingvvqamatter}, GQA \cite{hudson2019gqanewdatasetrealworld}, and Vibe-Eval \cite{padlewski2024vibeeval} contain some number-based questions, they are not systematically focused on challenging object enumeration. Thus, \textbf{CountQA} is explicitly designed to fill this void, providing the first dedicated benchmark to stress-test modern MLLMs on this fundamental task.
\section{The CountQA Dataset} \label{sec:data-creation}

\subsection{Data Collection and Curation}
\begin{figure}[t!]
\centering
\includegraphics[width=\columnwidth]{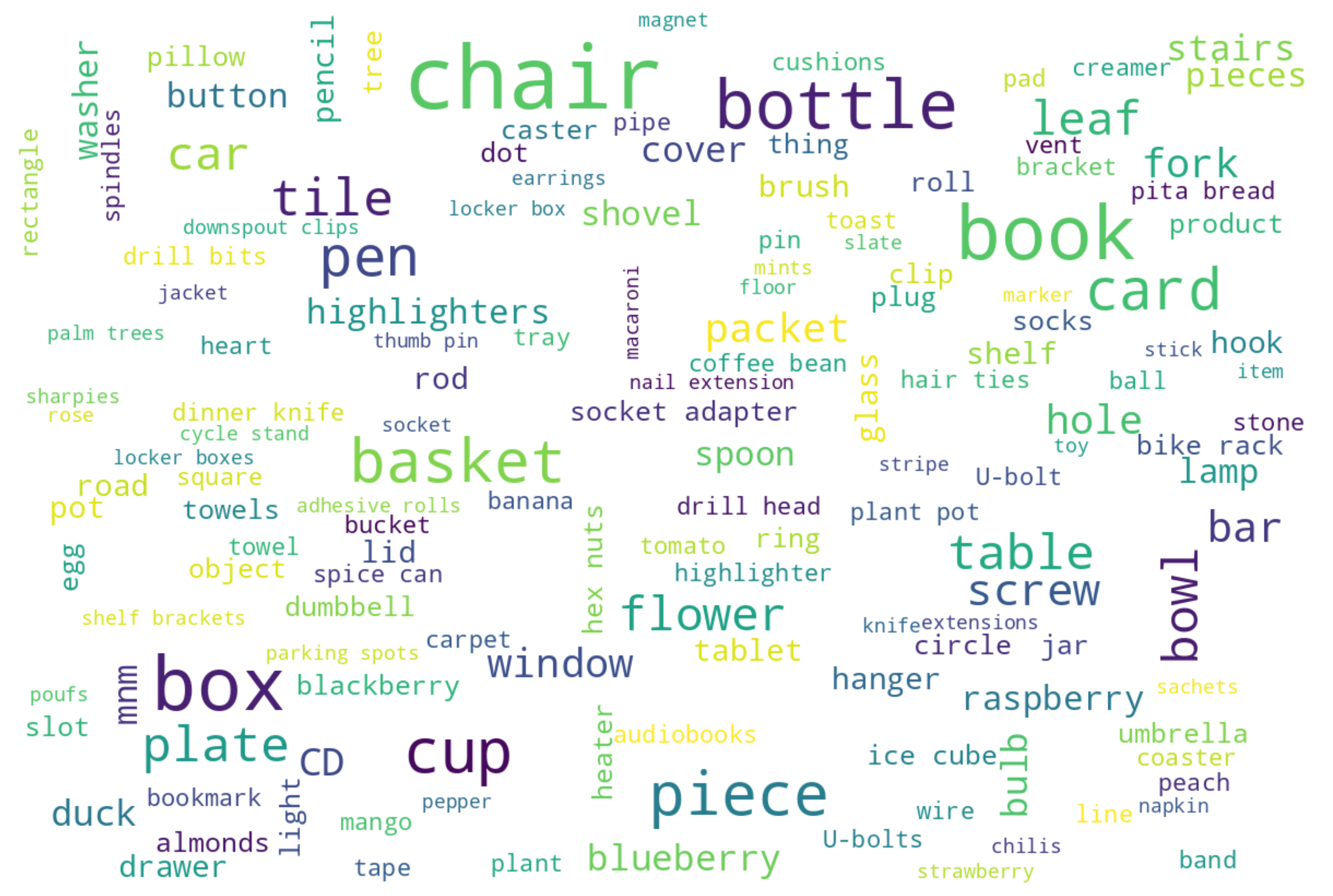}
\caption{A word cloud visualizing the frequency of objects in the CountQA dataset. The size of each word is directly proportional to the number of times that object appears in the questions.}
\label{fig:wordcloud}
\end{figure}

To create a dataset that serves as a robust proxy for real-world visual scenes, all images in CountQA were manually captured by the author(s). This process involved author(s) walking through various environments to capture a wide array of scenes and objects. The resulting collection contains a diverse mix of images, ranging from indoor domestic settings (e.g., items on a shelf, fruit in a bowl) to outdoor public spaces (e.g., cars in a parking lot, birds on a wire). This methodology ensures a departure from the iconic, and often over-represented, object classes found in large-scale web datasets, focusing instead on the ``long tail'' of everyday objects that MLLMs are more likely to encounter in real-world usage. Figure \ref{fig:wordcloud} visualizes the set of objects in our dataset.

A unique and critical aspect of our data collection protocol is that the ground truth (GT) count for each image was established \textit{in situ}, at the moment of image capture. Rather than relying on post-hoc annotation from a static 2D image, which can be ambiguous, the author-annotator could physically inspect the scene from multiple angles. This freedom of movement was crucial for resolving potential occlusions, verifying the identity of partially visible objects, and performing multiple recounts to establish a high-confidence ground truth. The annotators also performed additional checks by counting the objects in the image themselves to make sure that the objects were countable and the count matches the GT. This process inherently minimizes the annotation errors that can arise from the perceptual limitations of a single, fixed photograph, significantly bolstering the reliability of the GT labels in our dataset.

\subsection{QA Pair and Metadata Generation}
Following the initial data collection and in-situ ground truth (GT) counting, each image underwent a comprehensive annotation process to generate the final question-answer pairs and a rich set of metadata. The 1,528 QA pairs were generated through a semi-automated procedure where we used the Gemini 2.5 Flash model to formulate natural language questions based on the GT counts. This included both straightforward questions (e.g., ``How many eggs do you see?'') and more complex, compositional questions that require aggregating multiple object types (e.g., ``How many sharpies and pens do you see combined?''). The answers were directly inferred from the high-confidence GT counts.

\begin{figure*}[t!]
\centering
\includegraphics[width=\textwidth]{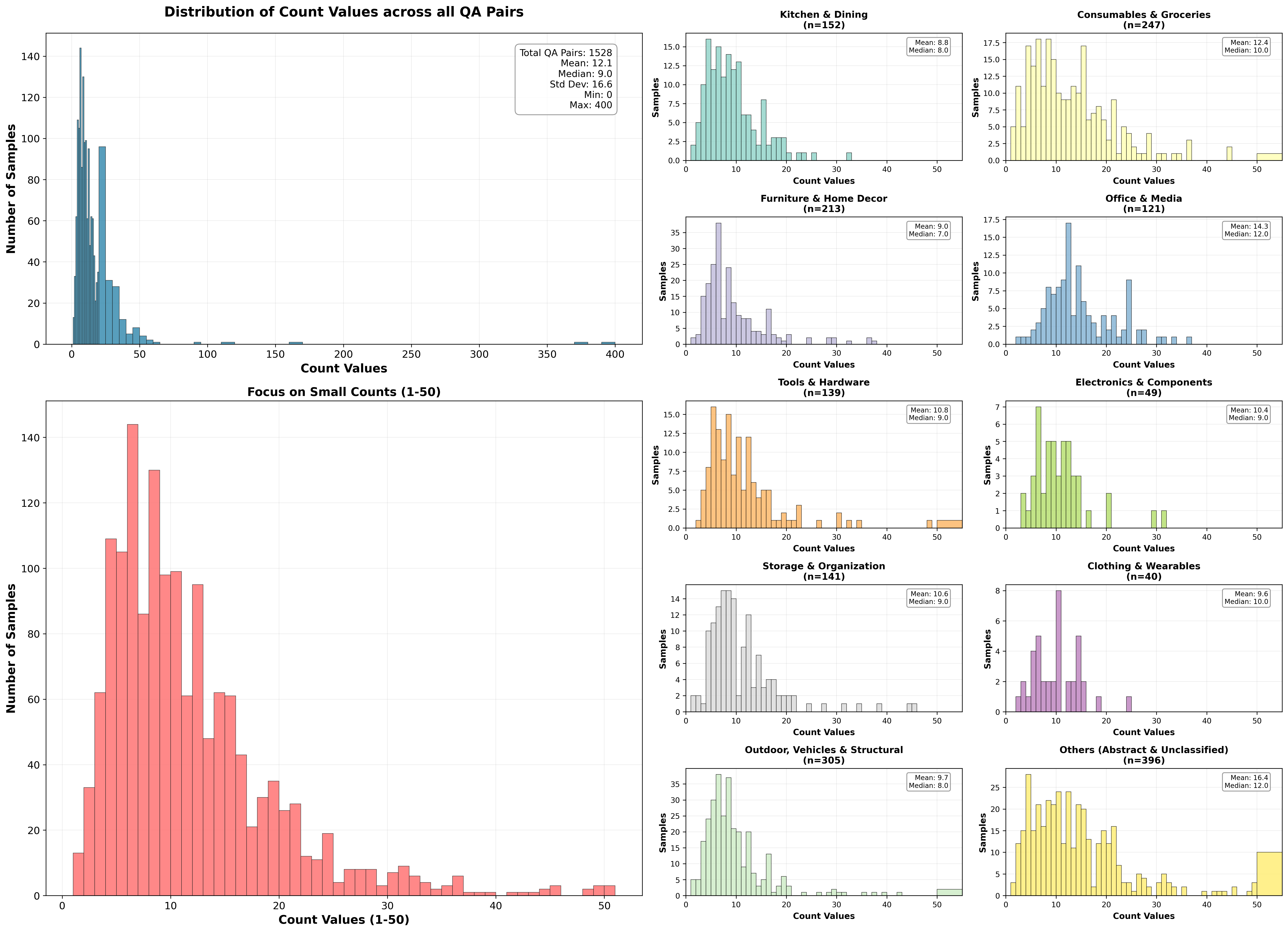}
\caption{Distribution of ground truth object counts in the CountQA dataset. 
\textbf{(Left)} The overall distribution across all 1,528 QA pairs. The histogram is right-skewed, showing a high frequency of low-count questions but also a long tail of challenging instances with high object counts. 
\textbf{(Right)} A breakdown of count distributions for each of the 10 dataset categories, revealing the varying numerical complexity across different domains.}
\label{fig:count_dist}
\end{figure*}

To facilitate detailed model evaluation, we further annotated the dataset with several layers of metadata. A complete list of all countable object types in each scene was programmatically extracted from the detailed notes taken by the author(s) during data collection. Furthermore, each image was manually assigned by the author(s) to one of ten predefined categories that cover a broad range of everyday contexts, from kitchen scenes to outdoor environments; the full list of these categories is detailed in the Figure \ref{fig:count_dist}. Finally, to explicitly measure model performance against visual noise, each image was manually labeled with an \texttt{is\_focused} tag. An image is considered `focused' if it features a clear subject with minimal background distractions, whereas `not focused' images are those that are visually busy, cluttered, or contain numerous objects competing for attention.

\begin{table}[t!]
\centering
\caption{Key statistics of the CountQA dataset.}
\label{tab:dataset_stats}
\begin{tabular}{lr}
\toprule
\textbf{Statistic} & \textbf{Value} \\
\midrule
\multicolumn{2}{c}{\textit{Overall Counts}} \\
\midrule
Images & 1,001 \\
QA Pairs & 1,528 \\
\midrule
\multicolumn{2}{c}{\textit{Vocabulary Statistics}} \\
\midrule
Vocab Size (Questions) & 524 \\
Avg. Question Length & 6.0 \\
\midrule
\multicolumn{2}{c}{\textit{Answer (GT) Statistics}} \\
\midrule
Min / Max Count & 0 / 400 \\
Mean / Median Count & 12.1 / 9 \\
Std. Dev. of Counts & 16.6 \\
\midrule
\multicolumn{2}{c}{\textit{Metadata Statistics}} \\
\midrule
Unique Categories & 10 \\
Unique objects & 441 \\
Focused Images & 524 \\
Cluttered Images & 477 \\
\bottomrule
\end{tabular}
\end{table}

\section{Experiments} \label{experiments}

\subsection{Models Evaluated}

Our evaluation focuses on prominent, general-purpose MLLMs rather than specialized counting frameworks. This choice is deliberate: the goal of CountQA is not to benchmark the state-of-the-art in specialized counting, but to probe the emergent capabilities of generalist models. We seek to answer whether precise, real-world object counting is a skill these powerful models possess out-of-the-box, so we restricted our evaluation to models that can be prompted with simple natural language questions.

We evaluated a total of 15 leading MLLMs, comprising both proprietary, closed-source models and a variety of prominent open-source alternatives. This diverse suite allows us to compare the performance of the largest, state-of-the-art systems with publicly accessible models that form the basis for much of academic research. A full list of all evaluated models and their versions is provided in our main results table.

\subsection{Evaluation Protocol}
\textbf{Task Framing and Prompting:} We frame the counting task as a regression problem, where the desired output for each question is a single integer. All models were evaluated in a zero-shot setting using a consistent system prompt designed to strictly constrain the output format. The prompt instructs the model to act as a counting assistant and to return its answer as a single integer and nothing else. To ensure deterministic and fact-based responses, the generation temperature for all models was set to 0. The full system prompt is provided in the Appendix \ref{sec:prompts} for reproducibility.

\textbf{Evaluation Metrics:} The primary metric for our evaluation is \textbf{Exact Match (EM)}, which measures the percentage of predictions where the final extracted number is identical to the ground truth integer. To provide a more nuanced view of performance, especially on questions with large counts, we also report on two \textbf{Relaxed Accuracy (RA)} metrics. An answer is considered correct under \textbf{RA@5\%} or \textbf{RA@10\%} if the predicted count falls within 5\% or 10\% of the ground truth value, respectively.

\textbf{Answer Parsing:} Our parsing strategy accounts for the varied adherence to the output format. For model outputs that were already a single integer, the value was used directly. However, several smaller open-source models frequently failed to adhere to the format, instead producing verbose text responses. To handle these cases, we employed a lenient, two-step parsing strategy: if an output was not a raw integer, we used Gemini 2.5 Flash as a rewriter LLM to extract the first numerical value from the verbose text. This post-processing step ensures that we can credit a model for identifying the correct count, even if it failed the formatting instruction.
\section{Results and Analysis} \label{results}

\begin{figure*}[t!]
\centering
\includegraphics[width=0.8\textwidth]{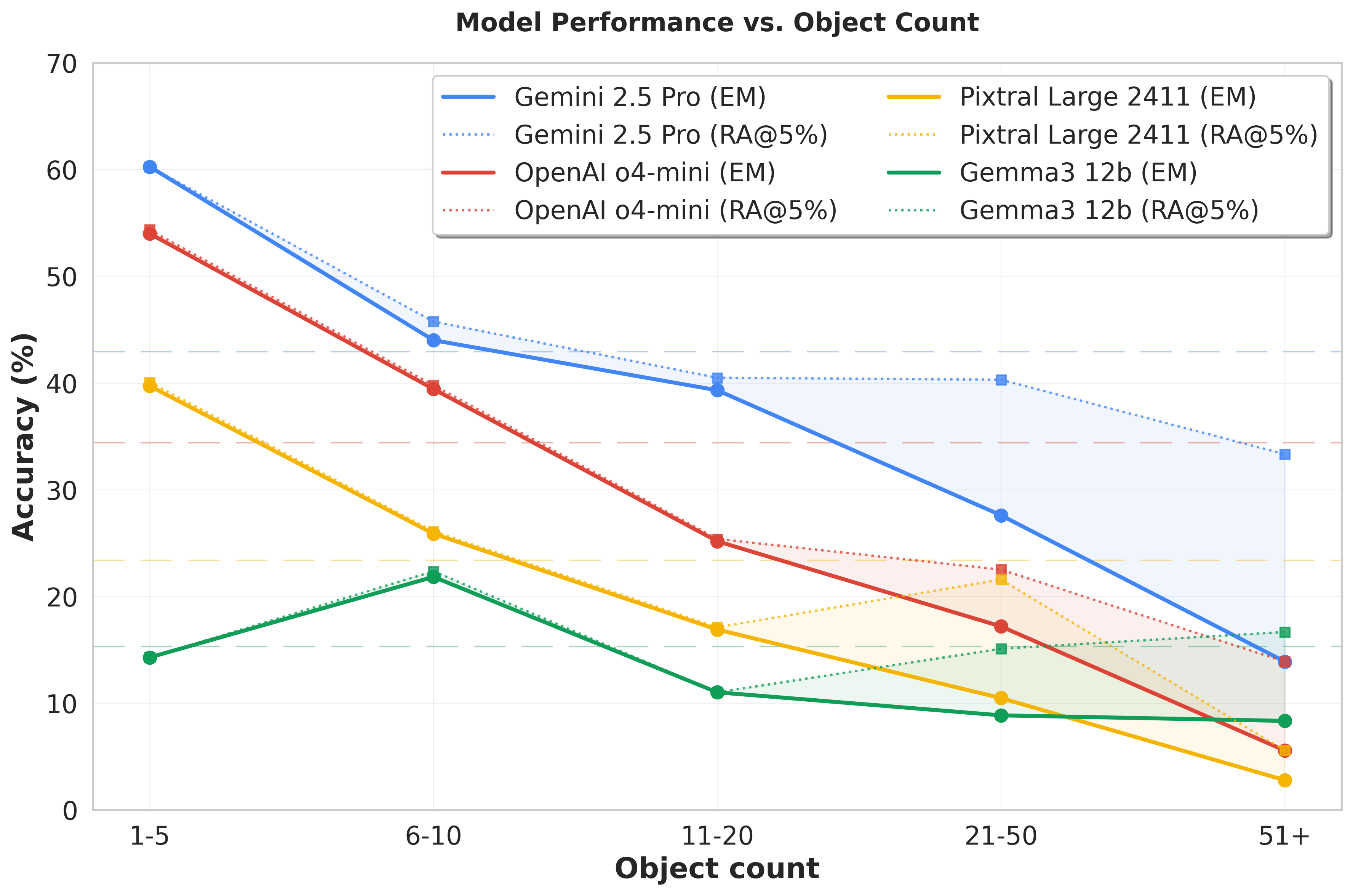}
\caption{Performance of four representative MLLMs on CountQA, analyzed by object count bins. The selected models represent a spectrum of performance and architecture from both proprietary and open-source classes.
\textbf{Solid lines} show Exact Match (EM) accuracy per bin; \textbf{dashed lines} indicate overall EM for each model; and the \textbf{shaded bands} show the margin of `near misses' up to the Relaxed Accuracy (RA@5\%) score.
The plot reveals several key findings. \textbf{Performance degrades} for all models as object counts increase. Even in the easiest bin (1-5 objects), the top model only reaches 60.3\% EM, highlighting the \textbf{dataset's difficulty}. For high counts (51+), accuracy \textbf{collapses} below 15\%. The widening bands suggest that models lose \textbf{precision} but can retain a general sense of quantity.}
\label{fig:performance_vs_counts}
\end{figure*}

\begin{figure}[t!]
\centering
\includegraphics[width=\columnwidth]{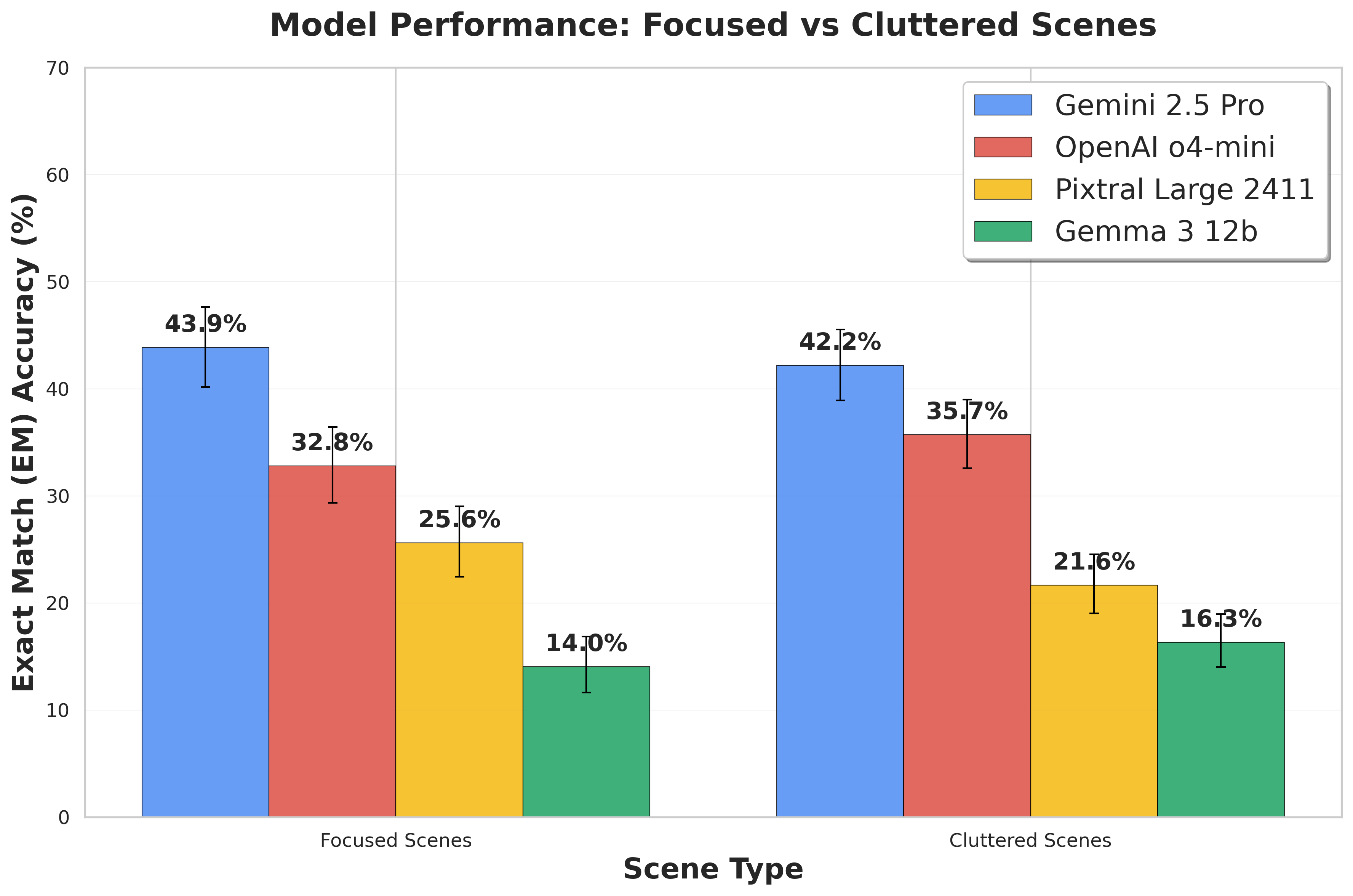}
\caption{Comparison of Exact Match (EM) accuracy on `Focused' versus `Cluttered' images for representative MLLMs. Each bar represents the EM accuracy for that subset, with error bars indicating the 95\% bootstrap confidence interval.}
\label{fig:focused_vs_cluttered}
\end{figure}

\begin{table*}[t!]
\centering
\caption{Performance of 15 MLLMs on the CountQA dataset, ranked by Exact Match (EM) accuracy. We also report Relaxed Accuracy at 5\% (RA@5\%) and 10\% (RA@10\%) thresholds. All scores are percentages (\%). The results highlight the challenging nature of the dataset, with the top-performing model achieving 42.93\% EM. A significant performance gap is observable between proprietary and open-source models. The best score in each column is in \textbf{bold}.}
\makebox[\textwidth][c]{
\label{tab:main_results}
\begin{tabular}{lcccc}
\toprule
\textbf{Model} & \textbf{Parameters} & \textbf{EM (\%) $\uparrow$} & \textbf{RA@5\% (\%) $\uparrow$} & \textbf{RA@10\% (\%) $\uparrow$} \\
\midrule
\multicolumn{5}{l}{\textit{Proprietary / API-based Models}} \\
\midrule
Gemini 2.5 Pro \cite{comanici2025gemini25pushingfrontier} & -- & \textbf{42.93 $\pm$ 2.48} & \textbf{45.99 $\pm$ 2.50} & \textbf{58.34 $\pm$ 2.47} \\
OpenAI o4-mini \cite{OpenAI2025o3} & -- & 34.42 $\pm$ 2.38 & 35.56 $\pm$ 2.40 & 45.71 $\pm$ 2.50 \\
OpenAI o3 \cite{OpenAI2025o3} & -- & 34.16 $\pm$ 2.38 & 36.13 $\pm$ 2.41 & 45.94 $\pm$ 2.49 \\
Gemini 2.5 Flash \cite{comanici2025gemini25pushingfrontier} & -- & 32.59 $\pm$ 2.35 & 36.80 $\pm$ 2.50 & 46.51 $\pm$ 2.58 \\
Claude 4 Sonnet \cite{Anthropic2025Claude4} & -- & 31.07 $\pm$ 2.32 & 32.39 $\pm$ 2.35 & 43.34 $\pm$ 2.49 \\
Qwen VL Max \cite{bai2025qwen25vltechnicalreport} & -- & 30.45 $\pm$ 2.31 & 31.96 $\pm$ 2.34 & 44.55 $\pm$ 2.49 \\
Mistral Medium \cite{Mistral2025Medium} & -- & 21.07 $\pm$ 2.04 & 24.74 $\pm$ 2.16 & 33.97 $\pm$ 2.37 \\
\midrule
\multicolumn{5}{l}{\textit{Open-Source Models}} \\
\midrule
Pixtral Large 2411 \cite{Mistral2024Pixtral} & 124B & 23.40 $\pm$ 2.12 & 25.07 $\pm$ 2.17 & 33.73 $\pm$ 2.37 \\
Mistral-small3.1-24b \cite{Mistral2025Small} & 24B & 20.75 $\pm$ 2.03 & 22.97 $\pm$ 2.11 & 32.40 $\pm$ 2.38 \\
Gemma3 27b \cite{gemmateam2025gemma3technicalreport} & 27B & 16.10 $\pm$ 1.84 & 16.95 $\pm$ 1.88 & 23.49 $\pm$ 2.12 \\
Gemma3 12b \cite{gemmateam2025gemma3technicalreport} & 12B & 15.31 $\pm$ 1.81 & 16.49 $\pm$ 1.86 & 22.19 $\pm$ 2.08 \\
Gemma3 4b \cite{gemmateam2025gemma3technicalreport} & 4B & 12.43 $\pm$ 1.66 & 12.83 $\pm$ 1.67 & 17.15 $\pm$ 1.89 \\
Llava 13b \cite{llava} & 13B & 10.93 $\pm$ 1.57 & 11.98 $\pm$ 1.67 & 15.36 $\pm$ 1.86 \\
Llava 34b \cite{llava} & 34B & 9.10 $\pm$ 1.44 & 9.23 $\pm$ 1.45 & 13.74 $\pm$ 1.73 \\
Llava 7b \cite{llava} & 7B & 8.37 $\pm$ 1.39 & 9.98 $\pm$ 1.63 & 13.89 $\pm$ 1.88 \\
\bottomrule
\end{tabular}
} 
\end{table*}

Our comprehensive evaluation of 15 leading Multimodal Large Language Models (MLLMs) on the CountQA benchmark demonstrates that direct, open-ended object counting remains a significant challenge for contemporary models. The results, summarized in Table~\ref{tab:main_results}, show a notable performance gap between proprietary and open-source models, as well as a sharp decline in accuracy as object counts increase. Refer to Figure \ref{fig:performance_vs_counts}.

\subsection{Overall Performance}

The results clearly indicate that even the most capable MLLMs struggle with precise enumeration. The top-performing model, Gemini 2.5 Pro, achieved an Exact Match (EM) accuracy of only 42.9\%. Following this, other leading proprietary models from OpenAI and Anthropic form a distinct performance tier, all scoring below 35\% EM.

A significant performance drop-off exists between these proprietary models and their open-source counterparts. The best-performing open-source model in our evaluation, Pixtral Large 2411, scored 23.4\% EM, which is nearly 20 percentage points lower than the leading model. This gap suggests that factors such as architectural scale, training data, or proprietary post-training methods provide a measurable, though ultimately insufficient, advantage for this task.

\subsection{The Impact of Object Count on Accuracy}

Across all MLLMs, accuracy degrades sharply as object counts grow. This finding aligns with existing literature, which notes that MLLM counting abilities are often brittle and break down as object counts increase beyond a small number \cite{qharabagh2025lvlmcountenhancingcountingability}. This trend is visualized in Figure~\ref{fig:performance_vs_counts} and a more detailed Table \ref{tab:appendix_perf_by_bin_long} in Appendix \ref{sec:Detailed results}.

\begin{itemize}
    \item \textbf{Small Counts (1-5 Objects):} In the range analogous to human ``subitizing'', models perform best. Gemini 2.5 Pro achieves 60.3\% EM, yet still fails on nearly 40\% of these simple prompts.
    \item \textbf{Moderate Counts (6-20 Objects):} Accuracy drops significantly once the object count exceeds the subitizing range. Gemini 2.5 Pro’s accuracy falls to 44.0\% for 6-10 objects and 39.3\% for 11-20 objects. This decline suggests a fundamental weakness in the models' ability to perform serial enumeration - the process of identifying and tallying individual items one by one.
    \item \textbf{High Counts (21+ Objects):} Performance collapses in higher-count bins. For scenes with over 50 objects, the top model's accuracy falls to just 13.9\%, and most other models are in the single digits. This demonstrates an almost complete failure to handle dense scenes.
\end{itemize}

The widening gap between Exact Match and Relaxed Accuracy at higher counts suggests that while models lose precision, they often retain an approximate sense of quantity. This points to a specific failure in the enumeration process rather than a complete loss of numerical understanding.

\subsection{The Role of Visual Clutter}

To isolate the effect of visual complexity, we tagged images as either `focused' or `cluttered'. As shown in Figure~\ref{fig:focused_vs_cluttered} and Table \ref{tab:appendix_perf_by_focus} in Appendix \ref{sec:Detailed results}, the results are nuanced. While some models performed worse on cluttered scenes as expected, several top-tier models, including OpenAI o4-mini and Qwen VL Max, paradoxically performed slightly better.

This counter-intuitive result is likely explained by a confounding variable: our `cluttered' scenes have a lower average object count (11.5) than our `focused' scenes (15.7). For the most advanced models, the reduced counting difficulty appears to outweigh the increased perceptual challenge of clutter.

\subsection{Architectural Bottlenecks in Fine-Grained Perception}

The systemic underperformance on our benchmark points to inherent architectural and training-related limitations in current MLLMs. Many successful approaches in the literature implicitly acknowledge this by developing "divide-and-conquer" systems \cite{qharabagh2025lvlmcountenhancingcountingability, shi2023trainingfreeobjectcountingprompts}. These frameworks, such as LVLM-Count, must break a single image into smaller patches to be counted individually, as the base model cannot reliably process the entire scene at once. The core problem appears to be a pipeline that is heavily optimized for semantic understanding at the cost of fine-grained perceptual accuracy. Several factors likely contribute to this:

\begin{itemize}
    \item \textbf{Lossy Modality Projection:} A common architectural pattern, seen in models like LLaVA, is to connect a vision encoder to the LLM with a simple linear projection layer \cite{llava}. This projection compresses a high-dimensional visual feature space into a sequence of tokens for the language model. This process is likely lossy, potentially discarding the precise spatial details required to distinguish one object from another in a dense arrangement.

    \item \textbf{Encoder Optimization Trade-offs:} The vision encoders themselves, often based on models like CLIP or SigLIP, are trained to align the overall semantics of an image with a text caption \cite{radford2021learningtransferablevisualmodels, zhai2023sigmoidlosslanguageimage}. This objective, focused on holistic semantic understanding, may not sufficiently prioritize the fine-grained visual acuity required for counting. The priority is capturing what the image is \emph{about}, not what it \emph{is composed of} at a granular level. Even in very large models like Qwen2-VL-72B and Pixtral Large, the vision component can be a relatively small, fixed-size encoder that may act as a bottleneck \cite{bai2025qwen25vltechnicalreport, Mistral2024Pixtral}.

    \item \textbf{Fixed-Resolution Processing:} Many vision encoders operate at a fixed, and often low, input resolution, which leads to a significant loss of detail in high-resolution images. While models like Gemma 3 use ``Pan \& Scan" techniques \cite{gemmateam2025gemma3technicalreport} and Qwen2-VL has introduced dynamic resolution mechanisms to address this \cite{bai2025qwen25vltechnicalreport}, it remains a common limitation. When an image with many small objects is downsampled to fit the encoder's input size, those objects can blur together or disappear entirely, making an accurate count impossible.
\end{itemize}

In essence, the architectural pipeline of typical MLLMs prioritizes a holistic, semantic interpretation of visual scenes over a detailed, object-centric parsing. Our work demonstrates that this architectural priority leads to a critical failure mode when models are tasked with precise enumeration. We perform an in-depth qualitative analysis in Appendix \ref{sec:qualitative analysis}.
\section{Conclusion and Future Directions} \label{conclusions}

This work introduced CountQA, a novel benchmark designed to assess the fundamental counting abilities of MLLMs. Our extensive evaluation of 15 models on its challenging real-world scenes reveals that direct object enumeration is a critical weakness for modern AI: the top-performing model achieved only 42.9\% accuracy, with performance degrading sharply as object counts increase. 

 Addressing this fundamental limitation points toward several promising avenues for future research:

\begin{itemize}
    \item \textbf{Novel Fusion Architectures:} Future work should explore new architectural designs that move beyond simple, lossy projection layers. Developing more sophisticated fusion mechanisms that can better preserve and integrate both the semantic and spatial details from the vision encoder is a critical next step.

    \item \textbf{Perception-Aware Training Objectives:} There is a need to develop pre-training or fine-tuning objectives for vision encoders that explicitly reward instance-level perception. While objectives like those used for CLIP are effective for semantic alignment, new tasks or loss functions could encourage the preservation of details necessary for distinguishing and enumerating discrete objects \cite{radford2021learningtransferablevisualmodels}.

    \item \textbf{Modular and Tool-Using MLLMs:} Instead of relying on a single end-to-end model, future systems could leverage the MLLM as a high-level reasoning engine that orchestrates specialized tools. Research into how MLLMs can effectively delegate fine-grained segmentation and localization to expert models, such as SAM, could yield more robust and reliable counting systems, as suggested by frameworks like LVLM-Count \cite{qharabagh2025lvlmcountenhancingcountingability}.

\end{itemize}

By focusing on these areas, the research community can work towards developing more perceptually grounded and numerically capable MLLMs, closing a critical gap in the capabilities of modern AI systems.

\nocite{yerramilli2024attributionregularizationmultimodalparadigms}
\nocite{yerramilli2024semanticaugmentationimagesusing}

\nocite{Jadhav_Cao_Shetty_Kumar_Sharma_Sukboontip_Tamarapalli_Zhang_Koul_2025}
\nocite{verma2024communitycentricperspectivecharacterizingdetecting}
\nocite{9412739}
\nocite{University2023}
\nocite{jain2023maeamultimodalattributionembodied}
\section{Limitations} \label{limitations}

While CountQA provides a robust new benchmark for evaluating the numerical reasoning of MLLMs, we acknowledge several limitations that define the scope of our contribution and offer avenues for future work.

First, the dataset's geographic and cultural scope is inherently linked to the environments of the author(s) who collected the images. While this grounds the dataset in real-world scenes, these scenes are primarily representative of specific residential and public areas. Consequently, the distribution of ``everyday objects'' may reflect a regional bias and may not encompass the full diversity of objects found in different global, cultural, or socio-economic contexts. Future work could build upon our methodology by curating similar high-quality, \textit{in situ} annotated data from a wider range of international locations.

Second, there is a deliberate trade-off in our dataset design between scale and annotation quality. The meticulous \textit{in situ} protocol for establishing ground truth counts is a core strength, as it minimizes the ambiguity and potential errors common in post-hoc annotation. However, this high-quality approach is labor-intensive and naturally limits the scale of the dataset compared to benchmarks created through large-scale, automated web scraping.

Finally, the scope of the questions in CountQA is focused primarily on direct enumeration (``How many X?'') and simple compositional counting (``How many X and Y?''). The dataset does not currently probe more complex forms of numerical or mathematical reasoning, such as comparative questions (``Are there more X's than Y's?''), parity judgments (``Is the number of X's even or odd?''), or other logical operations. These more complex reasoning tasks present a clear and exciting direction for future extensions of the dataset.
\section{Usage of Generative AI tools}
We utilized Generative AI tools to help improve the language, phrasing, and readability of this manuscript.
{
    \small
    \bibliographystyle{ieeenat_fullname}
    \bibliography{main}
}
\appendix

\onecolumn
\clearpage
\section{Detailed model performance}\label{sec:Detailed results}  
\begin{longtable}{llccc}
\caption{\textbf{Detailed model performance on the CountQA dataset, broken down by object count bins}. All scores are percentages (\%). Values are presented as `score ± margin', where the margin represents the 95\% bootstrap confidence interval.}
\label{tab:appendix_perf_by_bin_long} \\
\toprule
\textbf{Model} & \textbf{\# of objects} & \textbf{EM (\%) $\uparrow$} & \textbf{RA@5\% (\%) $\uparrow$} & \textbf{RA@10\% (\%) $\uparrow$} \\
\midrule
\endfirsthead
\caption[]{(continued)} \\
\toprule
\textbf{Model} & \textbf{Count Bin} & \textbf{EM (\%) $\uparrow$} & \textbf{RA@5\% (\%) $\uparrow$} & \textbf{RA@10\% (\%) $\uparrow$} \\
\midrule
\endhead
\bottomrule
\multicolumn{5}{r@{}}{\textit{Continued on next page}} \\
\endfoot
\bottomrule
\endlastfoot

\textbf{Gemini 2.5 Pro} & 1-5 & 60.28 $\pm$ 1.36 & 60.28 $\pm$ 1.36 & 63.07 $\pm$ 1.21 \\
& 6-10 & 44.01 $\pm$ 1.73 & 45.76 $\pm$ 1.71 & 52.47 $\pm$ 1.35 \\
& 11-20 & 39.33 $\pm$ 4.23 & 40.50 $\pm$ 4.26 & 64.03 $\pm$ 2.52 \\
& 21-50 & 27.60 $\pm$ 6.36 & 40.31 $\pm$ 6.37 & 53.93 $\pm$ 2.28 \\
& 51+ & 13.89 $\pm$ 14.72 & 33.33 $\pm$ 16.67 & 66.67 $\pm$ 2.78 \\
\midrule
\textbf{OpenAI o4-mini} & 1-5 & 54.01 $\pm$ 0.35 & 54.36 $\pm$ 0.35 & 55.05 $\pm$ 0.35 \\
& 6-10 & 39.44 $\pm$ 0.35 & 39.79 $\pm$ 0.35 & 44.89 $\pm$ 2.05 \\
& 11-20 & 25.17 $\pm$ 0.22 & 25.39 $\pm$ 0.22 & 44.27 $\pm$ 9.45 \\
& 21-50 & 17.19 $\pm$ 2.66 & 22.51 $\pm$ 2.66 & 39.79 $\pm$ 8.64 \\
& 51+ & 5.56 $\pm$ 8.33 & 13.89 $\pm$ 8.33 & 33.33 $\pm$ 2.78 \\
\midrule
\textbf{OpenAI o3} & 1-5 & 53.66 $\pm$ 0.35 & 53.66 $\pm$ 0.35 & 54.01 $\pm$ 0.35 \\
& 6-10 & 37.15 $\pm$ 2.38 & 37.15 $\pm$ 2.38 & 41.90 $\pm$ 2.38 \\
& 11-20 & 28.54 $\pm$ 0.22 & 28.76 $\pm$ 0.22 & 47.42 $\pm$ 9.33 \\
& 21-50 & 13.02 $\pm$ 5.21 & 23.44 $\pm$ 5.21 & 41.15 $\pm$ 8.85 \\
& 51+ & 13.89 $\pm$ 12.50 & 38.89 $\pm$ 13.89 & 52.78 $\pm$ 2.78 \\
\midrule
\textbf{Gemini 2.5 Flash} & 1-5 & 56.45 $\pm$ 1.35 & 57.80 $\pm$ 1.35 & 59.93 $\pm$ 1.35 \\
& 6-10 & 33.10 $\pm$ 1.45 & 34.55 $\pm$ 1.45 & 40.36 $\pm$ 2.81 \\
& 11-20 & 24.72 $\pm$ 2.87 & 27.59 $\pm$ 2.87 & 47.54 $\pm$ 10.00 \\
& 21-50 & 18.23 $\pm$ 7.55 & 33.33 $\pm$ 7.55 & 44.64 $\pm$ 5.62 \\
& 51+ & 8.33 $\pm$ 7.35 & 23.08 $\pm$ 7.35 & 26.92 $\pm$ 1.92 \\
\midrule
\textbf{Claude 4 Sonnet} & 1-5 & 44.91 $\pm$ 0.00 & 44.91 $\pm$ 0.00 & 46.67 $\pm$ 1.76 \\
& 6-10 & 34.81 $\pm$ 0.42 & 35.23 $\pm$ 0.42 & 38.97 $\pm$ 3.74 \\
& 11-20 & 22.35 $\pm$ 0.33 & 22.68 $\pm$ 0.33 & 45.80 $\pm$ 23.12 \\
& 21-50 & 24.48 $\pm$ 5.21 & 29.69 $\pm$ 5.21 & 44.27 $\pm$ 14.58 \\
& 51+ & 5.56 $\pm$ 16.67 & 22.22 $\pm$ 16.67 & 50.00 $\pm$ 27.78 \\
\midrule
\textbf{Qwen VL Max} & 1-5 & 46.69 $\pm$ 0.00 & 46.69 $\pm$ 0.00 & 48.43 $\pm$ 1.74 \\
& 6-10 & 35.51 $\pm$ 0.71 & 36.22 $\pm$ 0.71 & 40.46 $\pm$ 4.24 \\
& 11-20 & 22.57 $\pm$ 0.23 & 22.80 $\pm$ 0.23 & 49.66 $\pm$ 26.86 \\
& 21-50 & 14.58 $\pm$ 8.86 & 23.44 $\pm$ 8.86 & 43.23 $\pm$ 19.79 \\
& 51+ & 2.78 $\pm$ 2.78 & 5.56 $\pm$ 2.78 & 22.22 $\pm$ 16.67 \\
\midrule
\textbf{Mistral Medium} & 1-5 & 33.80 $\pm$ 6.62 & 40.42 $\pm$ 0.00 & 40.42 $\pm$ 0.00 \\
& 6-10 & 20.95 $\pm$ 1.76 & 22.71 $\pm$ 1.76 & 26.94 $\pm$ 4.23 \\
& 11-20 & 19.78 $\pm$ 2.02 & 21.80 $\pm$ 2.02 & 41.80 $\pm$ 20.00 \\
& 21-50 & 9.38 $\pm$ 8.85 & 18.23 $\pm$ 8.85 & 32.29 $\pm$ 14.06 \\
& 51+ & 0.00 $\pm$ 2.78 & 2.78 $\pm$ 2.78 & 5.56 $\pm$ 2.78 \\
\midrule
\textbf{Pixtral Large 2411} & 1-5 & 39.72 $\pm$ 0.35 & 40.07 $\pm$ 0.35 & 41.11 $\pm$ 1.04 \\
& 6-10 & 25.88 $\pm$ 0.22 & 26.10 $\pm$ 0.22 & 30.34 $\pm$ 4.24 \\
& 11-20 & 16.89 $\pm$ 0.23 & 17.12 $\pm$ 0.23 & 36.04 $\pm$ 18.92 \\
& 21-50 & 10.47 $\pm$ 11.11 & 21.58 $\pm$ 11.11 & 31.58 $\pm$ 10.00 \\
& 51+ & 2.78 $\pm$ 2.78 & 5.56 $\pm$ 2.78 & 11.11 $\pm$ 5.55 \\
\midrule
\textbf{Mistral-small3.1-24b} & 1-5 & 32.75 $\pm$ 0.00 & 32.75 $\pm$ 0.00 & 34.49 $\pm$ 1.74 \\
& 6-10 & 20.42 $\pm$ 2.12 & 22.54 $\pm$ 2.12 & 26.58 $\pm$ 4.04 \\
& 11-20 & 18.43 $\pm$ 0.22 & 18.65 $\pm$ 0.22 & 37.08 $\pm$ 18.43 \\
& 21-50 & 12.50 $\pm$ 8.33 & 20.83 $\pm$ 8.33 & 35.94 $\pm$ 15.11 \\
& 51+ & 2.78 $\pm$ 13.89 & 16.67 $\pm$ 13.89 & 30.56 $\pm$ 13.89 \\
\midrule
\textbf{Gemma3 27b} & 1-5 & 20.91 $\pm$ 0.34 & 21.25 $\pm$ 0.34 & 22.65 $\pm$ 1.40 \\
& 6-10 & 19.89 $\pm$ 0.53 & 20.42 $\pm$ 0.53 & 24.47 $\pm$ 4.05 \\
& 11-20 & 13.03 $\pm$ 0.23 & 13.26 $\pm$ 0.23 & 25.17 $\pm$ 11.91 \\
& 21-50 & 7.81 $\pm$ 4.17 & 11.98 $\pm$ 4.17 & 22.40 $\pm$ 10.42 \\
& 51+ & 0.00 $\pm$ 0.00 & 0.00 $\pm$ 0.00 & 0.00 $\pm$ 0.00 \\
\midrule
\textbf{Gemma3 12b} & 1-5 & 14.29 $\pm$ 0.00 & 14.29 $\pm$ 0.00 & 17.07 $\pm$ 2.78 \\
& 6-10 & 21.83 $\pm$ 0.53 & 22.36 $\pm$ 0.53 & 23.94 $\pm$ 1.58 \\
& 11-20 & 11.01 $\pm$ 0.00 & 11.01 $\pm$ 0.00 & 24.04 $\pm$ 13.03 \\
& 21-50 & 8.85 $\pm$ 6.25 & 15.10 $\pm$ 6.25 & 20.31 $\pm$ 5.21 \\
& 51+ & 8.33 $\pm$ 8.34 & 16.67 $\pm$ 8.34 & 22.22 $\pm$ 5.55 \\
\midrule
\textbf{Gemma3 4b} & 1-5 & 14.29 $\pm$ 0.00 & 14.29 $\pm$ 0.00 & 14.98 $\pm$ 0.69 \\
& 6-10 & 17.25 $\pm$ 0.18 & 17.43 $\pm$ 0.18 & 17.96 $\pm$ 0.53 \\
& 11-20 & 8.31 $\pm$ 0.00 & 8.31 $\pm$ 0.00 & 19.55 $\pm$ 11.24 \\
& 21-50 & 7.29 $\pm$ 2.61 & 9.90 $\pm$ 2.61 & 15.62 $\pm$ 5.72 \\
& 51+ & 0.00 $\pm$ 0.00 & 0.00 $\pm$ 0.00 & 0.00 $\pm$ 0.00 \\
\midrule
\textbf{Llava 13b} & 1-5 & 14.98 $\pm$ 0.27 & 15.25 $\pm$ 0.27 & 15.60 $\pm$ 0.35 \\
& 6-10 & 13.03 $\pm$ 1.26 & 14.29 $\pm$ 1.26 & 16.46 $\pm$ 2.17 \\
& 11-20 & 8.99 $\pm$ 0.86 & 9.85 $\pm$ 0.86 & 17.73 $\pm$ 7.88 \\
& 21-50 & 3.65 $\pm$ 1.49 & 5.14 $\pm$ 1.49 & 6.86 $\pm$ 1.72 \\
& 51+ & 8.33 $\pm$ 0.00 & 8.33 $\pm$ 0.00 & 11.11 $\pm$ 2.78 \\
\midrule
\textbf{Llava 34b} & 1-5 & 11.15 $\pm$ 0.00 & 11.15 $\pm$ 0.00 & 12.20 $\pm$ 1.05 \\
& 6-10 & 9.33 $\pm$ 0.00 & 9.33 $\pm$ 0.00 & 9.86 $\pm$ 0.53 \\
& 11-20 & 8.76 $\pm$ 0.00 & 8.76 $\pm$ 0.00 & 21.80 $\pm$ 13.04 \\
& 21-50 & 6.77 $\pm$ 1.04 & 7.81 $\pm$ 1.04 & 9.90 $\pm$ 2.09 \\
& 51+ & 5.56 $\pm$ 0.00 & 5.56 $\pm$ 0.00 & 8.33 $\pm$ 2.77 \\
\midrule
\textbf{Llava 7b} & 1-5 & 11.85 $\pm$ 1.38 & 13.23 $\pm$ 1.38 & 14.01 $\pm$ 0.78 \\
& 6-10 & 9.15 $\pm$ 1.15 & 10.30 $\pm$ 1.15 & 11.09 $\pm$ 0.79 \\
& 11-20 & 8.31 $\pm$ 1.40 & 9.71 $\pm$ 1.40 & 19.95 $\pm$ 10.24 \\
& 21-50 & 2.08 $\pm$ 1.97 & 4.05 $\pm$ 1.97 & 7.43 $\pm$ 3.38 \\
& 51+ & 2.78 $\pm$ 5.55 & 8.33 $\pm$ 5.55 & 16.67 $\pm$ 8.34 \\

\end{longtable}


\begin{longtable}{llccc}
\caption{\textbf{Detailed model performance on the CountQA dataset, broken down by `Focused' and `Cluttered' image subsets}. All scores are percentages (\%). Values are presented as `score ± margin`, where the margin represents the 95\% bootstrap confidence interval.}
\label{tab:appendix_perf_by_focus} \\
\toprule
\textbf{Model} & \textbf{Subset} & \textbf{EM (\%) $\uparrow$} & \textbf{RA@5\% (\%) $\uparrow$} & \textbf{RA@10\% (\%) $\uparrow$} \\
\midrule
\endfirsthead
\caption[]{(continued)} \\
\toprule
\textbf{Model} & \textbf{Subset} & \textbf{EM (\%) $\uparrow$} & \textbf{RA@5\% (\%) $\uparrow$} & \textbf{RA@10\% (\%) $\uparrow$} \\
\midrule
\endhead
\bottomrule
\multicolumn{5}{r@{}}{\textit{Continued on next page}} \\
\endfoot
\bottomrule
\endlastfoot

\textbf{Gemini 2.5 Pro} & Focused & 43.87 $\pm$ 3.73 & 47.03 $\pm$ 3.76 & 63.20 $\pm$ 3.63 \\
& Cluttered & 42.19 $\pm$ 3.31 & 45.17 $\pm$ 3.34 & 54.48 $\pm$ 3.34 \\
\midrule
\textbf{OpenAI o4-mini} & Focused & 32.79 $\pm$ 3.53 & 34.42 $\pm$ 3.57 & 47.56 $\pm$ 3.75 \\
& Cluttered & 35.72 $\pm$ 3.22 & 36.47 $\pm$ 3.23 & 44.24 $\pm$ 3.32 \\
\midrule
\textbf{OpenAI o3} & Focused & 33.68 $\pm$ 3.55 & 36.78 $\pm$ 3.72 & 50.07 $\pm$ 3.76 \\
& Cluttered & 34.55 $\pm$ 3.19 & 35.61 $\pm$ 3.23 & 42.66 $\pm$ 3.32 \\
\midrule
\textbf{Gemini 2.5 Flash} & Focused & 31.91 $\pm$ 3.51 & 36.98 $\pm$ 3.76 & 47.94 $\pm$ 3.89 \\
& Cluttered & 33.14 $\pm$ 3.16 & 36.66 $\pm$ 3.33 & 45.39 $\pm$ 3.44 \\
\midrule
\textbf{Claude 4 Sonnet} & Focused & 33.43 $\pm$ 3.56 & 34.93 $\pm$ 3.65 & 48.36 $\pm$ 3.77 \\
& Cluttered & 29.21 $\pm$ 3.05 & 30.38 $\pm$ 3.11 & 39.36 $\pm$ 3.28 \\
\midrule
\textbf{Qwen VL Max} & Focused & 28.49 $\pm$ 3.40 & 30.56 $\pm$ 3.47 & 45.85 $\pm$ 3.75 \\
& Cluttered & 32.00 $\pm$ 3.13 & 33.06 $\pm$ 3.16 & 43.53 $\pm$ 3.33 \\
\midrule
\textbf{Mistral Medium} & Focused & 21.12 $\pm$ 3.07 & 23.34 $\pm$ 3.17 & 35.60 $\pm$ 3.84 \\
& Cluttered & 21.03 $\pm$ 2.74 & 25.85 $\pm$ 2.87 & 32.67 $\pm$ 3.24 \\
\midrule
\textbf{Pixtral Large 2411} & Focused & 25.59 $\pm$ 3.28 & 27.37 $\pm$ 3.37 & 37.57 $\pm$ 3.64 \\
& Cluttered & 21.65 $\pm$ 2.77 & 23.23 $\pm$ 2.82 & 30.66 $\pm$ 3.09 \\
\midrule
\textbf{Mistral-small3.1-24b} & Focused & 19.50 $\pm$ 2.98 & 21.71 $\pm$ 3.13 & 31.31 $\pm$ 3.49 \\
& Cluttered & 21.74 $\pm$ 2.76 & 23.97 $\pm$ 2.82 & 33.25 $\pm$ 3.16 \\
\midrule
\textbf{Gemma3 27b} & Focused & 16.84 $\pm$ 2.82 & 17.87 $\pm$ 2.93 & 25.85 $\pm$ 3.32 \\
& Cluttered & 15.51 $\pm$ 2.44 & 16.22 $\pm$ 2.47 & 21.62 $\pm$ 2.65 \\
\midrule
\textbf{Gemma3 12b} & Focused & 14.03 $\pm$ 2.62 & 15.81 $\pm$ 2.73 & 23.04 $\pm$ 3.33 \\
& Cluttered & 16.33 $\pm$ 2.49 & 17.04 $\pm$ 2.54 & 21.50 $\pm$ 2.43 \\
\midrule
\textbf{Gemma3 4b} & Focused & 13.29 $\pm$ 2.56 & 13.59 $\pm$ 2.56 & 20.38 $\pm$ 3.24 \\
& Cluttered & 11.75 $\pm$ 2.17 & 12.22 $\pm$ 2.22 & 14.57 $\pm$ 2.20 \\
\midrule
\textbf{Llava 13b} & Focused & 10.34 $\pm$ 2.30 & 10.99 $\pm$ 2.38 & 15.88 $\pm$ 2.75 \\
& Cluttered & 11.40 $\pm$ 2.14 & 12.80 $\pm$ 2.31 & 14.93 $\pm$ 2.44 \\
\midrule
\textbf{Llava 34b} & Focused & 9.75 $\pm$ 2.24 & 9.90 $\pm$ 2.24 & 15.95 $\pm$ 2.78 \\
& Cluttered & 8.58 $\pm$ 1.89 & 8.70 $\pm$ 1.89 & 11.99 $\pm$ 2.14 \\
\midrule
\textbf{Llava 7b} & Focused & 8.86 $\pm$ 2.15 & 10.80 $\pm$ 2.34 & 16.55 $\pm$ 2.93 \\
& Cluttered & 7.99 $\pm$ 1.83 & 9.33 $\pm$ 2.08 & 11.80 $\pm$ 2.16 \\

\end{longtable}

\clearpage

\definecolor{tableheader}{gray}{0.85}
\definecolor{sectiongray}{gray}{0.95}

\definecolor{correctgreen}{rgb}{0.1, 0.5, 0.1}
\definecolor{wrongred}{rgb}{0.8, 0.1, 0.1}
\newcommand{\correct}[1]{{\color{correctgreen}#1}}
\newcommand{\wrong}[1]{{\color{wrongred}#1}}

\definecolor{questioncolor}{rgb}{0.1, 0.2, 0.7} 
\newcommand{\question}[1]{{\color{questioncolor}\textit{#1}}}
\newcommand{\gt}[1]{\textbf{#1}}

\begin{longtable}{p{6cm} p{4cm} m{3.5cm} m{1.5cm}}
\caption{Qualitative examples of model performance on the CountQA dataset. Each block shows an image, the question (in \question{blue italics}), the ground truth (in \gt{bold}), and the model outputs after rewrites (colored \correct{green} for correct and \wrong{red} for incorrect).} \\
\label{tab:qualitative_examples} \\
\toprule
\rowcolor{tableheader}
\textbf{Image} & \textbf{\begin{tabular}{@{}c@{}}Question \& \\ Ground Truth\end{tabular}} & \textbf{Model} & \textbf{Response} \\
\midrule
\endfirsthead
\caption[]{(continued)} \\
\toprule
\rowcolor{tableheader}
\textbf{Image} & \textbf{\begin{tabular}{@{}c@{}}Question \& \\ Ground Truth\end{tabular}} & \textbf{Model} & \textbf{Response} \\
\midrule
\endhead
\bottomrule
\endlastfoot


\rowcolor{sectiongray}
\multicolumn{4}{c}{\textbf{Section A: Simple Cases (Subitizing Range)}} \\
\midrule
\multirow{15}{*}{\includegraphics[width=4cm]{images/16.jpeg}} & 
\multirow{15}{*}{\parbox{4cm}{\question{How many marshmallows are there?} \newline \newline \gt{5}}} & 
Gemini 2.5 Pro & \correct{5} \\ \cdashline{3-4}
& & OpenAI o4-mini & \wrong{4} \\ \cdashline{3-4}
& & OpenAI o3 & \correct{5} \\ \cdashline{3-4}
& & Gemini 2.5 Flash & \correct{5} \\ \cdashline{3-4}
& & Claude 4 Sonnet & \wrong{6} \\ \cdashline{3-4}
& & Qwen VL Max & \wrong{6} \\ \cdashline{3-4}
& & Mistral Medium & \wrong{6} \\ \cdashline{3-4}
& & Pixtral Large 2411 & \wrong{6} \\ \cdashline{3-4}
& & Mistral-small3.1-24b & \wrong{6} \\ \cdashline{3-4}
& & Gemma3 27b & \wrong{6} \\ \cdashline{3-4}
& & Gemma3 12b & \wrong{6} \\ \cdashline{3-4}
& & Gemma3 4b & \wrong{6} \\ \cdashline{3-4}
& & Llava 13b & \wrong{4} \\ \cdashline{3-4}
& & Llava 34b & \wrong{6} \\ \cdashline{3-4}
& & Llava 7b & \wrong{6} \\
\midrule
\multirow{15}{*}{\includegraphics[width=4cm]{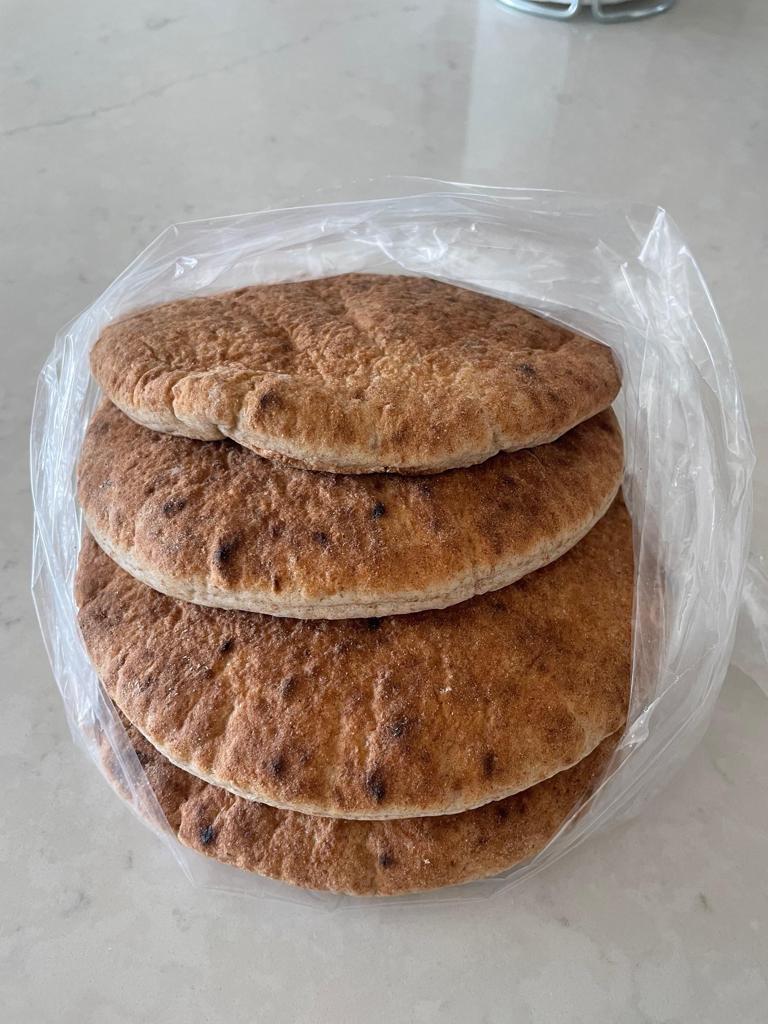}} & 
\multirow{15}{*}{\parbox{4cm}{\question{How many pita breads are there?} \newline \newline \gt{4}}} & 
Gemini 2.5 Pro & \correct{4} \\ \cdashline{3-4}
& & OpenAI o4-mini & \correct{4} \\ \cdashline{3-4}
& & OpenAI o3 & \correct{4} \\ \cdashline{3-4}
& & Gemini 2.5 Flash & \correct{4} \\ \cdashline{3-4}
& & Claude 4 Sonnet & \correct{4} \\ \cdashline{3-4}
& & Qwen VL Max & \correct{4} \\ \cdashline{3-4}
& & Mistral Medium & \correct{4} \\ \cdashline{3-4}
& & Pixtral Large 2411 & \correct{4} \\ \cdashline{3-4}
& & Mistral-small3.1-24b & \correct{4} \\ \cdashline{3-4}
& & Gemma3 27b & \correct{4} \\ \cdashline{3-4}
& & Gemma3 12b & \wrong{6} \\ \cdashline{3-4}
& & Gemma3 4b & \correct{4} \\ \cdashline{3-4}
& & Llava 13b & \correct{4} \\ \cdashline{3-4}
& & Llava 34b & \wrong{5} \\ \cdashline{3-4}
& & Llava 7b & \correct{4} \\
\midrule
\rowcolor{sectiongray}
\multicolumn{4}{c}{\textbf{Section B: Counting Many Items}} \\
\midrule
\multirow{15}{*}{\includegraphics[width=4cm]{images/32.jpeg}} & 
\multirow{15}{*}{\parbox{4cm}{\question{How many MnMs are there?} \newline \newline \gt{21}}} & 
Gemini 2.5 Pro & \wrong{18} \\ \cdashline{3-4}
& & OpenAI o4-mini & \wrong{16} \\ \cdashline{3-4}
& & OpenAI o3 & \wrong{18} \\ \cdashline{3-4}
& & Gemini 2.5 Flash & \correct{21} \\ \cdashline{3-4}
& & Claude 4 Sonnet & \wrong{20} \\ \cdashline{3-4}
& & Qwen VL Max & \wrong{20} \\ \cdashline{3-4}
& & Mistral Medium & \wrong{19} \\ \cdashline{3-4}
& & Pixtral Large 2411 & \wrong{22} \\ \cdashline{3-4}
& & Mistral-small3.1-24b & \wrong{24} \\ \cdashline{3-4}
& & Gemma3 27b & \correct{21} \\ \cdashline{3-4}
& & Gemma3 12b & \wrong{20} \\ \cdashline{3-4}
& & Gemma3 4b & \wrong{8} \\ \cdashline{3-4}
& & Llava 13b & \wrong{0} \\ \cdashline{3-4}
& & Llava 34b & \wrong{15} \\ \cdashline{3-4}
& & Llava 7b & \wrong{12} \\
\midrule
\multirow{15}{*}{\includegraphics[width=4cm]{images/4.jpeg}} & 
\multirow{15}{*}{\parbox{4cm}{\question{How many leaves are there?} \newline \newline \gt{8}}} & 
Gemini 2.5 Pro & \wrong{11} \\ \cdashline{3-4}
& & OpenAI o4-mini & \correct{8} \\ \cdashline{3-4}
& & OpenAI o3 & \wrong{7} \\ \cdashline{3-4}
& & Gemini 2.5 Flash & \wrong{10} \\ \cdashline{3-4}
& & Claude 4 Sonnet & \correct{8} \\ \cdashline{3-4}
& & Qwen VL Max & \wrong{7} \\ \cdashline{3-4}
& & Mistral Medium & \wrong{12} \\ \cdashline{3-4}
& & Pixtral Large 2411 & \wrong{13} \\ \cdashline{3-4}
& & Mistral-small3.1-24b & \wrong{10} \\ \cdashline{3-4}
& & Gemma3 27b & \wrong{11} \\ \cdashline{3-4}
& & Gemma3 12b & \wrong{12} \\ \cdashline{3-4}
& & Gemma3 4b & \wrong{22} \\ \cdashline{3-4}
& & Llava 13b & \wrong{15} \\ \cdashline{3-4}
& & Llava 34b & \wrong{10} \\ \cdashline{3-4}
& & Llava 7b & \correct{8} \\
\midrule
\multirow{15}{*}{\includegraphics[width=4cm]{images/37.jpeg}} & 
\multirow{15}{*}{\parbox{4cm}{\question{How many almonds are there?} \newline \newline \gt{14}}} & 
Gemini 2.5 Pro & \wrong{15} \\ \cdashline{3-4}
& & OpenAI o4-mini & \wrong{13} \\ \cdashline{3-4}
& & OpenAI o3 & \correct{14} \\ \cdashline{3-4}
& & Gemini 2.5 Flash & \wrong{13} \\ \cdashline{3-4}
& & Claude 4 Sonnet & \wrong{18} \\ \cdashline{3-4}
& & Qwen VL Max & \wrong{13} \\ \cdashline{3-4}
& & Mistral Medium & \wrong{11} \\ \cdashline{3-4}
& & Pixtral Large 2411 & \wrong{13} \\ \cdashline{3-4}
& & Mistral-small3.1-24b & \correct{14} \\ \cdashline{3-4}
& & Gemma3 27b & \wrong{17} \\ \cdashline{3-4}
& & Gemma3 12b & \wrong{18} \\ \cdashline{3-4}
& & Gemma3 4b & \wrong{13} \\ \cdashline{3-4}
& & Llava 13b & \wrong{9} \\ \cdashline{3-4}
& & Llava 34b & \wrong{15} \\ \cdashline{3-4}
& & Llava 7b & \wrong{12} \\
\midrule
\multirow{15}{*}{\includegraphics[width=4cm]{images/280.jpeg}} & 
\multirow{15}{*}{\parbox{4cm}{\question{How many bananas and kinderjoys are there?} \newline \newline \gt{11}}} & 
Gemini 2.5 Pro & \correct{11} \\ \cdashline{3-4}
& & OpenAI o4-mini & \wrong{12} \\ \cdashline{3-4}
& & OpenAI o3 & \wrong{10} \\ \cdashline{3-4}
& & Gemini 2.5 Flash & \correct{11} \\ \cdashline{3-4}
& & Claude 4 Sonnet & \correct{11} \\ \cdashline{3-4}
& & Qwen VL Max & \wrong{10} \\ \cdashline{3-4}
& & Mistral Medium & \wrong{9} \\ \cdashline{3-4}
& & Pixtral Large 2411 & \wrong{8} \\ \cdashline{3-4}
& & Mistral-small3.1-24b & \wrong{12} \\ \cdashline{3-4}
& & Gemma3 27b & \wrong{10} \\ \cdashline{3-4}
& & Gemma3 12b & \wrong{7} \\ \cdashline{3-4}
& & Gemma3 4b & \correct{11} \\ \cdashline{3-4}
& & Llava 13b & \wrong{10} \\ \cdashline{3-4}
& & Llava 34b & \wrong{9} \\ \cdashline{3-4}
& & Llava 7b & \wrong{9} \\
\midrule
\rowcolor{sectiongray}
\multicolumn{4}{c}{\textbf{Section C: Performance on Cluttered Scenes}} \\
\midrule
\multirow{15}{*}{\includegraphics[width=4cm]{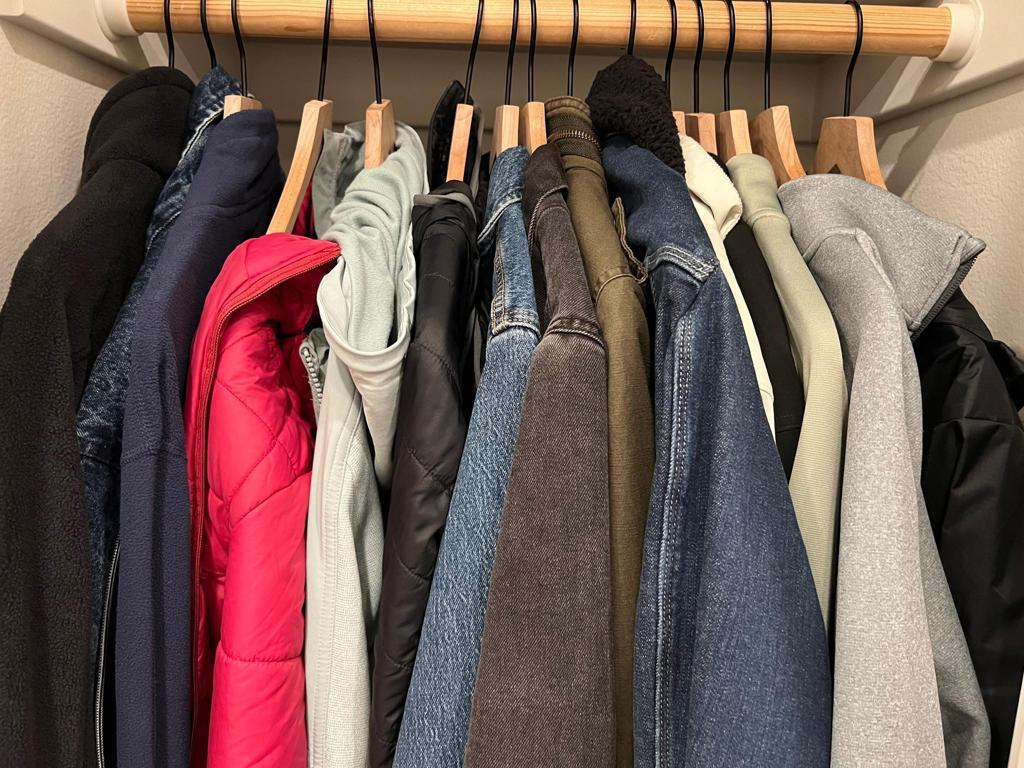}} & 
\multirow{15}{*}{\parbox{4cm}{\question{How many jackets are there?} \newline \newline \gt{15}}} & 
Gemini 2.5 Pro & \wrong{14} \\ \cdashline{3-4}
& & OpenAI o4-mini & \wrong{13} \\ \cdashline{3-4}
& & OpenAI o3 & \wrong{12} \\ \cdashline{3-4}
& & Gemini 2.5 Flash & \wrong{11} \\ \cdashline{3-4}
& & Claude 4 Sonnet & \wrong{4} \\ \cdashline{3-4}
& & Qwen VL Max & \wrong{10} \\ \cdashline{3-4}
& & Mistral Medium & \wrong{11} \\ \cdashline{3-4}
& & Pixtral Large 2411 & \wrong{10} \\ \cdashline{3-4}
& & Mistral-small3.1-24b & \wrong{11} \\ \cdashline{3-4}
& & Gemma3 27b & \wrong{3} \\ \cdashline{3-4}
& & Gemma3 12b & \wrong{6} \\ \cdashline{3-4}
& & Gemma3 4b & \wrong{3} \\ \cdashline{3-4}
& & Llava 13b & \wrong{12} \\ \cdashline{3-4}
& & Llava 34b & \correct{15} \\ \cdashline{3-4}
& & Llava 7b & \wrong{12} \\
\midrule
\multirow{15}{*}{\includegraphics[width=4cm]{images/2.jpeg}} & 
\multirow{15}{*}{\parbox{4cm}{\question{How many products are there?} \newline \newline \gt{21}}} & 
Gemini 2.5 Pro & \correct{21} \\ \cdashline{3-4}
& & OpenAI o4-mini & \wrong{19} \\ \cdashline{3-4}
& & OpenAI o3 & \correct{21} \\ \cdashline{3-4}
& & Gemini 2.5 Flash & \wrong{20} \\ \cdashline{3-4}
& & Claude 4 Sonnet & \wrong{23} \\ \cdashline{3-4}
& & Qwen VL Max & \wrong{18} \\ \cdashline{3-4}
& & Mistral Medium & \wrong{19} \\ \cdashline{3-4}
& & Pixtral Large 2411 & \wrong{19} \\ \cdashline{3-4}
& & Mistral-small3.1-24b & \wrong{23} \\ \cdashline{3-4}
& & Gemma3 27b & \correct{21} \\ \cdashline{3-4}
& & Gemma3 12b & \wrong{30} \\ \cdashline{3-4}
& & Gemma3 4b & \wrong{16} \\ \cdashline{3-4}
& & Llava 13b & \wrong{Invalid Response} \\ \cdashline{3-4}
& & Llava 34b & \wrong{25} \\ \cdashline{3-4}
& & Llava 7b & \wrong{13} \\
\midrule
\rowcolor{sectiongray}
\multicolumn{4}{c}{\textbf{Section D: Challenges in Object Recognition}} \\
\midrule
\multirow{15}{*}{\includegraphics[width=4cm]{images/209.jpeg}} & 
\multirow{15}{*}{\parbox{4cm}{\question{How many tennis balls are there?} \newline \newline \gt{5}}} & 
Gemini 2.5 Pro & \correct{5} \\ \cdashline{3-4}
& & OpenAI o4-mini & \correct{5} \\ \cdashline{3-4}
& & OpenAI o3 & \correct{5} \\ \cdashline{3-4}
& & Gemini 2.5 Flash & \correct{5} \\ \cdashline{3-4}
& & Claude 4 Sonnet & \correct{5} \\ \cdashline{3-4}
& & Qwen VL Max & \correct{5} \\ \cdashline{3-4}
& & Mistral Medium & \correct{5} \\ \cdashline{3-4}
& & Pixtral Large 2411 & \correct{5} \\ \cdashline{3-4}
& & Mistral-small3.1-24b & \wrong{4} \\ \cdashline{3-4}
& & Gemma3 27b & \wrong{7} \\ \cdashline{3-4}
& & Gemma3 12b & \wrong{6} \\ \cdashline{3-4}
& & Gemma3 4b & \wrong{7} \\ \cdashline{3-4}
& & Llava 13b & \wrong{6} \\ \cdashline{3-4}
& & Llava 34b & \wrong{7} \\ \cdashline{3-4}
& & Llava 7b & \wrong{6} \\
\midrule
\multirow{15}{*}{\includegraphics[width=4cm]{images/212.jpeg}} & 
\multirow{15}{*}{\parbox{4cm}{\question{How many plates are there?} \newline \newline \gt{4}}} & 
Gemini 2.5 Pro & \correct{4} \\ \cdashline{3-4}
& & OpenAI o4-mini & \wrong{5} \\ \cdashline{3-4}
& & OpenAI o3 & \wrong{5} \\ \cdashline{3-4}
& & Gemini 2.5 Flash & \correct{4} \\ \cdashline{3-4}
& & Claude 4 Sonnet & \correct{4} \\ \cdashline{3-4}
& & Qwen VL Max & \wrong{5} \\ \cdashline{3-4}
& & Mistral Medium & \wrong{5} \\ \cdashline{3-4}
& & Pixtral Large 2411 & \correct{4} \\ \cdashline{3-4}
& & Mistral-small3.1-24b & \wrong{7} \\ \cdashline{3-4}
& & Gemma3 27b & \wrong{6} \\ \cdashline{3-4}
& & Gemma3 12b & \wrong{9} \\ \cdashline{3-4}
& & Gemma3 4b & \wrong{6} \\ \cdashline{3-4}
& & Llava 13b & \correct{4} \\ \cdashline{3-4}
& & Llava 34b & \wrong{10} \\ \cdashline{3-4}
& & Llava 7b & \wrong{0} \\
\midrule
\rowcolor{sectiongray}
\multicolumn{4}{c}{\textbf{Section E: Challenges in Attribute Recognition}} \\
\midrule
\multirow{15}{*}{\includegraphics[width=4cm]{images/639.jpeg}} & 
\multirow{15}{*}{\parbox{4cm}{\question{How many white hexagons are there?} \newline \newline \gt{9}}} & 
Gemini 2.5 Pro & \correct{9} \\ \cdashline{3-4}
& & OpenAI o4-mini & \wrong{8} \\ \cdashline{3-4}
& & OpenAI o3 & \correct{9} \\ \cdashline{3-4}
& & Gemini 2.5 Flash & \correct{9} \\ \cdashline{3-4}
& & Claude 4 Sonnet & \wrong{8} \\ \cdashline{3-4}
& & Qwen VL Max & \wrong{10} \\ \cdashline{3-4}
& & Mistral Medium & \wrong{10} \\ \cdashline{3-4}
& & Pixtral Large 2411 & \wrong{8} \\ \cdashline{3-4}
& & Mistral-small3.1-24b & \wrong{11} \\ \cdashline{3-4}
& & Gemma3 27b & \wrong{7} \\ \cdashline{3-4}
& & Gemma3 12b & \wrong{6} \\ \cdashline{3-4}
& & Gemma3 4b & \wrong{12} \\ \cdashline{3-4}
& & Llava 13b & \wrong{6} \\ \cdashline{3-4}
& & Llava 34b & \wrong{10} \\ \cdashline{3-4}
& & Llava 7b & \wrong{6} \\
\midrule
\multirow{15}{*}{\includegraphics[width=4cm]{images/569.jpeg}} & 
\multirow{15}{*}{\parbox{4cm}{\question{How many red thumb pins are there?} \newline \newline \gt{5}}} & 
Gemini 2.5 Pro & \correct{5} \\ \cdashline{3-4}
& & OpenAI o4-mini & \correct{5} \\ \cdashline{3-4}
& & OpenAI o3 & \wrong{6} \\ \cdashline{3-4}
& & Gemini 2.5 Flash & \wrong{7} \\ \cdashline{3-4}
& & Claude 4 Sonnet & \wrong{4} \\ \cdashline{3-4}
& & Qwen VL Max & \correct{5} \\ \cdashline{3-4}
& & Mistral Medium & \wrong{6} \\ \cdashline{3-4}
& & Pixtral Large 2411 & \wrong{4} \\ \cdashline{3-4}
& & Mistral-small3.1-24b & \wrong{6} \\ \cdashline{3-4}
& & Gemma3 27b & \wrong{11} \\ \cdashline{3-4}
& & Gemma3 12b & \wrong{8} \\ \cdashline{3-4}
& & Gemma3 4b & \wrong{6} \\ \cdashline{3-4}
& & Llava 13b & \wrong{10} \\ \cdashline{3-4}
& & Llava 34b & \wrong{10} \\ \cdashline{3-4}
& & Llava 7b & \wrong{10} \\
\end{longtable}

\twocolumn

\section{Qualitative Analysis}\label{sec:qualitative analysis}

To better understand the nature of model failures on \textbf{CountQA}, we present a qualitative analysis of representative examples, with full model outputs available in Appendix Table~\ref{tab:qualitative_examples}. These examples reveal that the quantitative scores are a result of several distinct failure modes, ranging from simple inaccuracies to more profound limitations in perception and reasoning.

\textbf{Performance on Simple Scenes.}
We first examine scenes with a small number of objects (fewer than 10), which should be trivial for most models. In cases like counting four pita breads or five marshmallows, we observe that most high-performing proprietary models provide the correct answer. However, even in this simple regime, errors are surprisingly common. For the marshmallow example, a significant portion of the models, including several powerful open-source systems, produce \textit{off-by-one} errors. This suggests that even at a scale where humans can subitize instantly, a fundamental lack of precision affects many current MLLMs.

\textbf{The Challenge of Counting Many Items.}
As the number of objects increases, model performance degrades significantly, corroborating our quantitative findings. In examples with moderate counts, such as counting a handful of M\&Ms or almonds, we observe a wide variance in errors.  For instance, when tasked with counting a pile of 14 almonds, the best models provide answers near the ground truth, such as Gemini 2.5 Pro's prediction of 15. In sharp contrast, other models exhibit a significant failure of enumeration on the exact same image, with Llava 13b predicting only 9. This demonstrates that beyond a small threshold, the counting ability of some models does not just degrade gracefully with slight inaccuracies, but can collapse entirely.Furthermore, compositional questions requiring the summation of multiple object types, such as ``bananas and kinderjoys,'' prove challenging even for top models.

\textbf{Impact of Clutter and Scene Complexity.}
Our analysis of cluttered scenes reveals a complex interaction between perceptual challenge and model capability. In an example requiring the count of 15 jackets in a cluttered pile, nearly every model fails, most by a significant margin. Interestingly, this example also highlights the unpredictability of performance, as \texttt{Llava 34b} - a model with a low overall EM score - is the only one to correctly identify the count, while models that rank much higher fail. This suggests that certain visual configurations can coincidentally align with a model's specific biases. However, the general trend points to clutter being a major source of error, with some models failing completely, such as \texttt{Llava 13b} providing an invalid response when tasked with counting products on a cluttered shelf.

\textbf{Failures in Recognition and Grounding.}
Perhaps the most insightful failures are those that stem not from counting itself, but from the preceding step of visual perception. In several cases, it appears the models fail because they cannot correctly identify the objects they are asked to count, for instance when asked to count specific `plates' from a stack. This issue is even more pronounced when \textbf{attribute binding} is required. When asked to count ``how many \textit{white} hexagons'' from a tiled surface or ``\textit{red} thumb pins'' on a board, the error rates are extremely high across almost all models. This indicates a significant failure in compositional visual grounding - the ability to filter objects based on multiple properties (e.g., color + shape + object class) before performing the enumeration. This suggests that the core challenge often lies in correctly parsing the visual scene according to the prompt's constraints, long before the numerical part of the task begins.

\section{MLLM Prompts}\label{sec:prompts}

All MLLMs were provided the following system prompt along with the question in user prompt to perform the counting task. 
\begin{tcolorbox}[title=System Prompt]
``You are a helpful assistant that counts the number of items in an image. The user will provide an image and ask a question about the number of a certain type of item in the image. If the user question is referring to multiple objects, it means that you need to provide a sum of the number of items. You will count the number of items and return the number as an integer. Your output should STRICTLY be a single integer and nothing else."
\end{tcolorbox}

\end{document}